\documentclass[10pt,letterpaper]{article}

\usepackage[final]{cogsci}

\usepackage{graphicx}      
\usepackage{booktabs}      
\usepackage{caption}       
\usepackage{subcaption}    
\usepackage{needspace}     
\usepackage{hyperref}      

\usepackage{amsmath,amssymb}

\hypersetup{
    colorlinks=true,
    linkcolor=blue,
    citecolor=blue,
    urlcolor=blue
}

\newcommand{\datasetname}{\textsc{FlipSet}} 

\usepackage[utf8]{inputenc}

\usepackage{microtype}

\usepackage{inconsolata}

\usepackage{graphicx}

\usepackage[nameinlink,capitalise,noabbrev]{cleveref}

\usepackage[
  style=apa,
  natbib=true,
  annotation=false,
]{biblatex}
\usepackage{float} 

\title{Egocentric Bias in Vision-Language Models}

\author[1]{\mbox{Maijunxian Wang}\thanks{Equal Contributions}\footnote[3]}
\author[2]{\mbox{Yijiang Li}\footnote[1]}
\author[3]{\mbox{Bingyang Wang}}
\author[4]{\mbox{Tianwei Zhao}}
\author[5]{\mbox{Ran Ji}}
\author[6]{\mbox{Qingying Gao}\thanks{Equal Advising}}
\author[7]{\mbox{Emmy Liu}\footnote[2]}
\author[8]{\mbox{Hokin Deng}\footnote[2]}
\author[9]{\mbox{Dezhi Luo}\thanks{Correspondence: mjxwang@berkeley.edu; ihzedoul@umich.edu}}

\affil[1]{Cognitive Science Program, University of California, Berkeley}
\affil[2]{Department of Electrical and Computer Engineering, University of California San Diego}
\affil[3]{School of Computer Science, Georgia Institute of Technology \& Emory University}
\affil[4]{Department of Computer Science, Johns Hopkins University}
\affil[5]{Department of Cognitive Science, University of California San Diego}
\affil[6]{Department of Computer Science \& Wilmer Eye Institute, Johns Hopkins University}
\affil[7]{Language Technologies Institute, Carnegie Mellon University}
\affil[8]{Robotics Institute, Carnegie Mellon University}
\affil[9]{Weinberg Institute for Cognitive Science, University of Michigan}

\begin{document}

\maketitle

\begin{abstract}
Visual perspective taking-inferring how the world appears from another's viewpoint-is foundational to social cognition. We introduce FlipSet, a diagnostic benchmark for Level-2 visual perspective taking (L2 VPT) in vision-language models. The task requires simulating 180-degree rotations of 2D character strings from another agent's perspective, isolating spatial transformation from 3D scene complexity. Evaluating 103 VLMs reveals systematic egocentric bias: the vast majority perform below chance, with roughly three-quarters of errors reproducing the camera viewpoint. Control experiments expose a compositional deficit-models achieve high theory-of-mind accuracy and above-chance mental rotation in isolation, yet fail catastrophically when integration is required. This dissociation indicates that current VLMs lack the mechanisms needed to bind social awareness to spatial operations, suggesting fundamental limitations in model-based spatial reasoning. FlipSet provides a cognitively grounded testbed for diagnosing perspective-taking capabilities in multimodal systems.

\end{abstract}

\section{Introduction}

Visual perspective taking (VPT), the ability to reason about how a scene appears from another's point of view, is foundational to both human social cognition and situated artificial intelligence. It underlies skills such as interpreting spatial instructions, anticipating occluded views, and understanding what others can or cannot see. Cognitive science distinguishes between two levels of this ability. Level-1 (L1) VPT involves recognizing whether an object is visible from a particular viewpoint, while Level-2 (L2) VPT concerns how that object appears—for example, understanding that a "6" might look like a "9" from across the table \citep{zhao2016nine}. Unlike L1, L2 VPT requires mentally transforming spatial representations, often through mental rotation or embodied simulation \citep{shepard1971mental, gallese1998mirror}, and typically emerges later in human development \citep{edwards2019level, Moll2011look}. The development of L2 VPT, described by Jean Piaget \citep{piaget1954construction} as "the elimination of egocentrism," is regarded as a developmental landmark in children's social cognition \citep{flavell2013perspectives}.

Recent advances in Vision-Language Models (VLMs) \citep{llavavideo, radford2021learning, alayrac2022flamingo, li2023blip2} have demonstrated remarkable capabilities in perception \citep{wang2024qwen2vlenhancingvisionlanguagemodels, chen2025expandingperformanceboundariesopensource, jiang2025videop2r, gemmateam2025gemma3technicalreport, cheng2024videollama2advancingspatialtemporal}, reasoning \citep{zhang2024improve, xu2024llava, cheng2024vision}, and visual commonsense understanding \citep{zellers2019recognitioncognitionvisualcommonsense, park2020visualcometreasoningdynamiccontext}. These advancements prompt the considerations of whether VLMs are capable of situated social reasoning that necessitates taking the perspective of another agent when it conflicts with their own. In this respectm, L2 VPT poses a particularly stringent benchmark, testing whether models can go beyond line-of-sight reasoning to simulate another's visual experience—a capability essential for effective interaction in real-world environments.

In this paper, we introduce \datasetname, a diagnostic benchmark that isolates the spatial transformation component of L2 VPT while controlling for basic theory-of-mind demands. Each item shows a printed card with 2D symbols (e.g., "81") and a plush monkey on the opposite side, facing the card's back. The model is asked: "What does the monkey see on the card?" To answer correctly, models must mentally rotate the card 180°, simulating the monkey's perspective. By using 2D strings that transform under rotation rather than complex 3D scenes, we minimize confounding spatial complexity. Crucially, we introduce control conditions that explicitly separate theory-of-mind recognition (understanding that another's view differs) from mental rotation capabilities (transforming spatial representations), enabling precise diagnosis of where models fail in the L2 VPT cognitive pipeline.

We evaluate 103 publicly available VLMs under unified zero-shot conditions
, analyzing not only accuracy but the distribution of error types through systematic answer choice design: correct, egocentric (camera viewpoint), confusable (visually similar), and random. We find that 91.3\% of models perform below the 25\% chance level, with 75.88\% of errors being egocentric—models simply reproduce what the camera sees. Chain-of-Thought prompting fails to mitigate and often amplifies this bias. However, control experiments on 24 models reveal a more nuanced picture: models achieve high theory-of-mind accuracy (90.4\%), recognizing that agents see differently, yet perform near chance on isolated mental rotation (26.1\%) and catastrophically on L2 VPT (10.3\%). Critically, most models show L2 VPT performance below what their component abilities predict, revealing a compositional deficit—models possess cognitive building blocks but cannot integrate them in situated reasoning contexts.

\textbf{Our contributions are threefold:}
\begin{itemize}
    \item We introduce \datasetname, a controlled benchmark for L2 VPT using 2D rotation that isolates spatial transformation from 3D complexity and theory-of-mind recognition from mental rotation capabilities, enabling the first large-scale evaluation (103 VLMs) that can precisely diagnose component failures.
    \item Through systematic answer choice design distinguishing correct, egocentric, confusable, and random responses, we reveal dominant egocentric bias (75.88\% of errors).
    \item Via controlled experiments separating theory-of-mind, mental rotation, and L2 VPT, we provide behavioral reductionistic evidence of a compositional deficit: models recognize perspective differences (ToM: 90.4\%) and perform above-chance mental rotation in isolation (MR: 26.1\%), yet fail catastrophically when integration is required (L2 VPT: 10.3\%), illuminating fundamental limitations in current VLM architecture.
\end{itemize}

\section{Related Work}

\paragraph{Human Cognitive Science.}

VPT plays a foundational role in social cognition and spatial reasoning. Cognitive science distinguishes between L1 VPT—judging what is visible from another's viewpoint—and L2 VPT—judging how an object appears from that viewpoint. Crucially, L2 VPT is a composite ability that integrates two distinct cognitive requirements: it demands not only basic theory of mind understanding that others perceive the world differently, but also the capacity to perform structured mental operations that actively transform spatial representations \citep{Moll2011look, edwards2019level, flavell2013perspectives}. This dual requirement situates L2 VPT within Piaget's Concrete Operational Stage \citep{piaget1954construction, piaget1977development}, the developmental period during which children acquire the ability to mentally manipulate spatial relationships through reversible, systematic operations. The mental transformation component involves model-based simulation rather than static perception, as evidenced by classic studies showing that humans incur a linear reaction time cost proportional to the degree of rotation required \citep{shepard1971mental, tarr1989mental}. Neuroimaging studies further demonstrate that L2 perspective taking activates the intraparietal sulcus and premotor cortex—regions associated with embodied simulation of spatial transformations and another person's position \citep{zacks2008neuroimaging, gallese2007before, gunia2021brain}. This composite architecture—requiring both perspective awareness and spatial transformation capabilities—distinguishes L2 VPT from simpler visibility judgments and makes it a principled probe for embodied social understanding in artificial systems.

\vspace{-3mm}

\paragraph{Cognitive Benchmarking in VLMs.}

Recent benchmarks have attempted to assess VPT in VLMs but suffer from confounding factors that prevent clear diagnostic interpretation. CLEVR-PT \citep{singh2023spatialvlm} and Spatial-VQA \citep{li2024spatialvqa} introduce multi-camera or 3D setups that conflate viewpoint simulation with depth perception, object occlusion, and multi-object tracking. PerspectBench \citep{gao2024vision} (part of CoreCognition \citep{li2025core}), COMFORT \citep{zhang2024vision}, and Omni-Perspective \citep{wang2025vision} assess L2 VPT using controlled and real-life 3D spatial arrangements that are primarily inspired by Piaget's three-mountain task \citep{piaget1977development}, where children must describe how a scene of three mountains appears from different viewpoints. However, precisely because L2 VPT comprises both theory of mind understanding and spatial transformation abilities, benchmarks relying on complex 3D configurations introduce a critical diagnostic ambiguity: when models fail, we cannot determine whether the deficit stems from inability to recognize that others see differently (ToM), inability to mentally transform spatial representations (mental rotation), or both. To address this limitation, we design \datasetname{} to isolate the spatial transformation component while controlling for basic ToM demands. Our benchmark uses simple two-dimensional symbols that appear different under 180-degree rotation, minimizing spatial complexity as a confounding factor, and introduces control conditions that explicitly separate ToM recognition from mental rotation capabilities, enabling precise diagnosis of where models fail in the L2 VPT cognitive pipeline.



\begin{figure*}[!ht]
  \centering
  
  \begin{subfigure}[b]{0.48\textwidth}
  \centering
  \includegraphics[width=\textwidth]{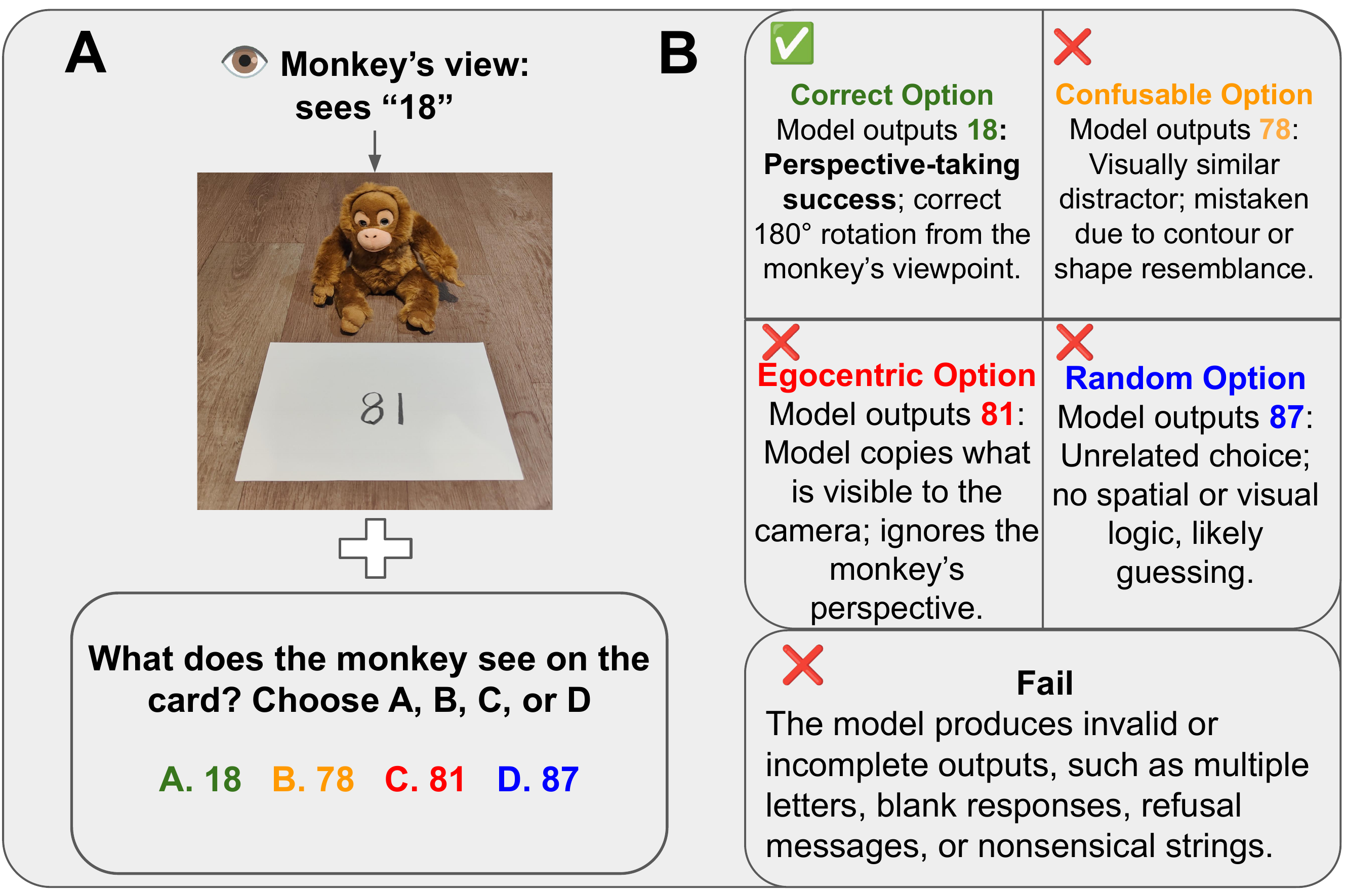}
  \caption{Main Experiments}

  \label{fig:flipset-example}
  \end{subfigure}
  \begin{subfigure}[b]{0.48\textwidth}
  \centering
  \includegraphics[width=\textwidth]{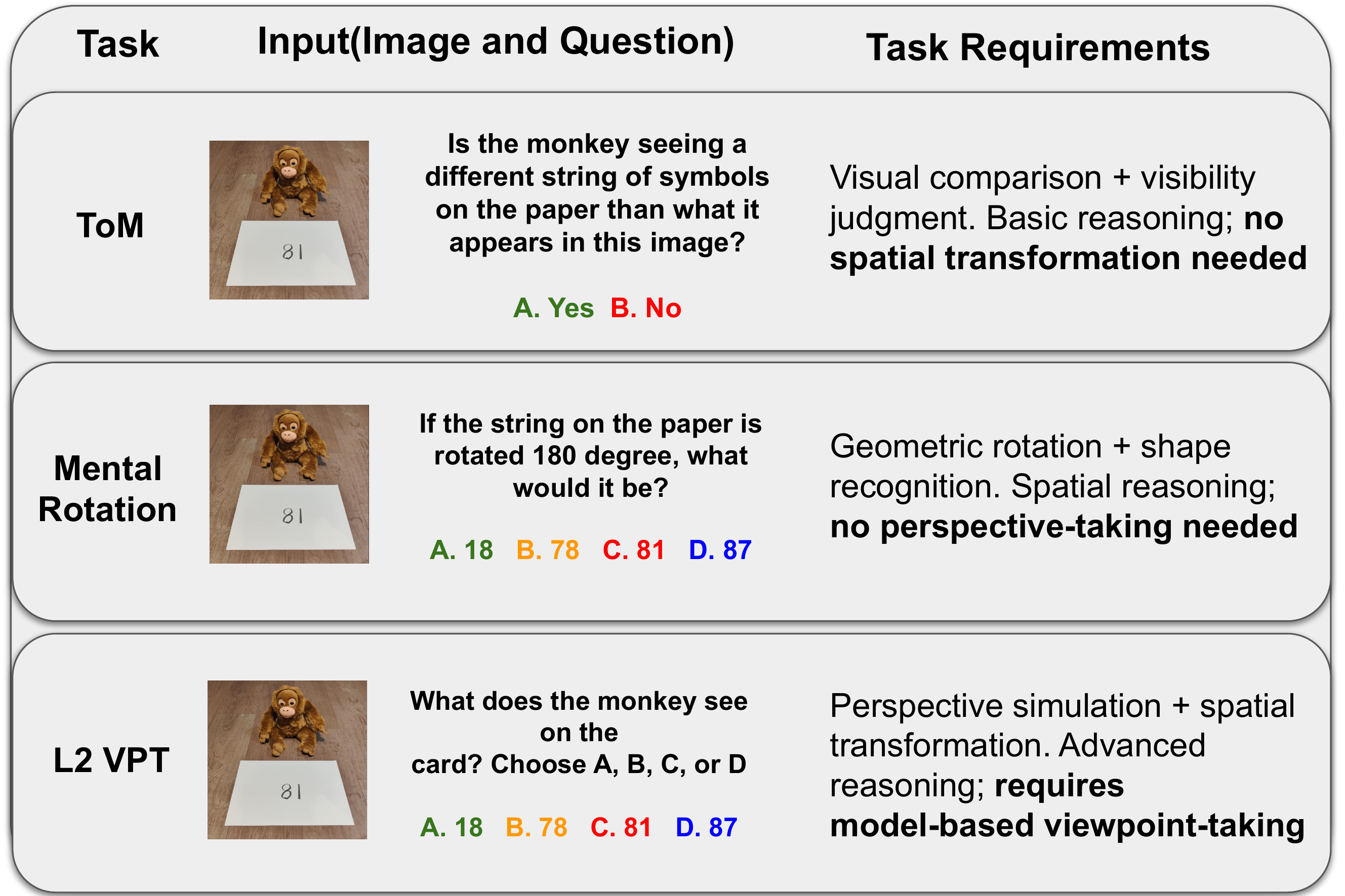}
  
  \caption{Control Experiment}
  \label{fig:task_comparison}
  \end{subfigure}
\caption{\textit{FlipSet benchmark design and evaluation approach.}
\textbf{(a)} \textbf{Prompt type and Error types in model responses across cognitive tasks.} Each FlipSet item asks the model what a monkey sees on the back of a card—requiring a 180° mental rotation from the monkey's viewpoint. Answer options correspond to distinct reasoning outcomes: 
\textcolor{green!60!black}{Correct} (successful perspective taking), 
\textcolor{red!70!black}{Egocentric} (front-view repetition), 
\textcolor{orange!90!black}{Confusable} (visually similar distractor), 
\textcolor{blue!70!black}{Random} (unrelated guess), and 
\textcolor{gray!60!black}{Fail} (invalid or empty output).  
\textbf{(b)} \textbf{Three cognitive tasks comparison.} L1 VPT requires simple visibility judgment, MR involves pure geometric transformation, and L2 VPT demands complex perspective simulation. All tasks use identical visual stimuli.}

\vspace{-2mm}

  \label{fig:flipset-main}
\end{figure*}

\section{Methods}
\label{sec:benchmarkD}

\subsection{Main Experiments}

We introduce \datasetname, a diagnostic benchmark designed to evaluate L2 VPT in VLMs. Building on the seminal 6$\rightarrow$9 rotation paradigm introduced by \citet{zhao2016nine}, we expand this approach into a systematic benchmark that isolates perspective taking from confounding spatial complexity. Unlike the three-mountain paradigm and its computational variants that rely on complex 3D spatial arrangements \citep{piaget1977development, gao2024vision}, we use 2D strings that transform under 180° rotation (e.g., "81"$\rightarrow$"18", "pond"$\rightarrow$"puod"). This design reduces cognitive load from depth perception, occlusion, and object tracking, enabling us to strictly isolate the mental rotation component of L2 VPT from broader 3D spatial understanding. 

Each \datasetname{} item consists of an image paired with a multiple-choice question, as illustrated in \autoref{fig:flipset-example}. The image shows a white A4 card (21$\times$30 cm) with centrally printed black Arial characters, placed upright on a wooden floor. The camera faces the card front-on, while a plush monkey sits on the opposite side, facing the card's back. To answer correctly, models must mentally rotate the card 180° to simulate the monkey's viewpoint, requiring genuine perspective transformation rather than surface pattern recognition.

\datasetname{} contains 28 baseline items organized by difficulty, ranging from simple digit reversals to complex alphanumeric transformations (\Cref{tab:questionset}). Items progress from symmetric digit pairs (e.g., "81"$\rightarrow$"18") to letter mirroring (e.g., "nod"$\rightarrow$"pou") and mixed alphanumeric rearrangement (e.g., "W819"$\rightarrow$"618M"). All strings use Arabic numerals and English alphabet letters—a constraint that ensures most VLMs have substantial training exposure to these characters while enabling tight experimental control. All visual elements—font size, lighting, camera angle, and background—are standardized to minimize perceptual artifacts.

\begin{table}[t]

\small\centering

\begin{tabular}{llc}

\toprule

Category & Example & \#Items\\

\midrule

Two Digits               & 81 $\!\rightarrow\!$ 18  & 2\\

Two Digits (mixed)       & 89 $\!\rightarrow\!$ 68  & 2\\

Three Digits (mixed)     & 168 $\!\rightarrow\!$ 891 & 4\\

Single Letter            & d  $\!\rightarrow\!$ p   & 4\\

Two Letters              & dd $\!\rightarrow\!$ pp  & 2\\

Two Letters (mixed)      & do $\!\rightarrow\!$ op  & 4\\

Three Letters (mixed)    & nod $\!\rightarrow\!$ pou & 6\\

Four Letters (mixed)     & pond $\!\rightarrow\!$ puod & 2\\

Four Combo (mixed)       & W819 $\!\rightarrow\!$ 618M & 2\\

\bottomrule

\end{tabular}

\caption{\datasetname{} question classes ordered by difficulty (28 items total).}

\vspace{-5mm}

\label{tab:questionset}

\end{table}

Crucially, we extend beyond prior work through systematic answer choice design that enables diagnostic error analysis. Each item includes four answer choices reflecting distinct cognitive strategies: correct (perspective-transformed answer), egocentric (camera viewpoint), confusable (visually similar distractor), and random (unrelated option). For example, when the card shows "81" from the camera view, the choices are "18" (correct), "81" (egocentric), "78" (confusable), and "87" (random), as shown in \autoref{fig:flipset-example}. To eliminate positional bias, we systematically permute these four choices across 12 counterbalanced layouts per item
, yielding 28 items $\times$ 12 layouts = 336 evaluation instances per model. Analysis confirms minimal positional effects across layouts
. We classify all model responses into five categories that reveal distinct failure modes beyond binary accuracy:

\begin{itemize}

    \item \textbf{Correct} — Outputs "18," successfully simulating the rotated viewpoint through mental transformation.

    \item \textbf{Egocentric} — Outputs "81," reproducing the camera view and failing to adopt the monkey's perspective.

    \item \textbf{Confusable} — Selects "78," showing partial visual reasoning based on shape similarity to the correct answer without completing the perspective transformation.

    \item \textbf{Random} — Outputs "87," an unrelated option with no plausible geometric relation, indicating arbitrary guessing.

    \item \textbf{Fail} — Produces invalid, empty, or nonsensical outputs (e.g., "ABC" or "I don't know"), reflecting instruction-following failures rather than visual reasoning.

\end{itemize}

\begin{table*}[t]
\centering
\scriptsize
\setlength{\tabcolsep}{3pt}
\begin{tabular}{l|cccccccccccc}
\toprule
 & \textbf{Llama} & \textbf{LLaVA} & \textbf{BLIP} & \textbf{InternVL} & \textbf{Video} & \textbf{DeepSeek} & \textbf{InternLM} & \textbf{Molmo} & \textbf{VILA} & \textbf{Gemma} & \textbf{Janus} & \textbf{Qwen} \\
\midrule
\# Models & 15 & 12 & 8 & 6 & 5 & 5 & 5 & 4 & 4 & 4 & 4 & 3 \\
Params & 7-90B & 2-72B & 3-13B & 1-38B & 7-13B & 3-28B & 7-8B & 1-72B & 3-40B & 3-27B & 1-7B & 3-72B \\
Acc (\%) & 4.8 & 9.3 & 13.8 & 12.0 & 10.6 & 4.5 & 1.4 & 6.1 & 8.6 & 13.2 & 4.5 & 17.3 \\
Ego (\%) & 88.4 & 72.6 & 62.2 & 70.8 & 57.0 & 93.9 & 97.3 & 79.2 & 80.7 & 75.5 & 91.6 & 53.9 \\
\midrule
 & \textbf{Parrot} & \textbf{Moondream} & \textbf{Yi-VL} & \textbf{ChatUniVi} & \textbf{EMU} & \textbf{MiniGPT} & \textbf{MiniMonkey} & \textbf{XVERSE} & \textbf{WeMM} & \textbf{VisualGLM} & \textbf{TransCore} & \textbf{Falcon} \\
\midrule
\# Models & 2 & 2 & 2 & 2 & 2 & 1 & 1 & 1 & 1 & 1 & 1 & 1 \\
Params & 7-14B & 2B & 6-34B & 8-13B & 8-37B & 13B & 2B & 13B & 7B & 6B & 13B & 11B \\
Acc (\%) & 1.5 & 3.9 & 0.4 & 17.6 & 2.2 & 19.0 & 0.9 & 22.9 & 9.5 & 6.0 & 33.0 & 25.0 \\
Ego (\%) & 91.7 & 87.5 & 98.5 & 60.7 & 90.3 & 41.4 & 97.3 & 24.2 & 87.5 & 34.8 & 33.3 & 50.0 \\
\midrule
 & \textbf{Pixtral} & \textbf{Phi} & \textbf{Fuyu} & \textbf{Other} & \textbf{OpenFlam.} & \textbf{OmniLMM} & \textbf{NVLM} & \textbf{GLM} & \textbf{Monkey} & \textbf{MGM} & \textbf{360VL} & \\
\midrule
\# Models & 1 & 1 & 1 & 1 & 1 & 1 & 1 & 1 & 1 & 1 & 1 & \\
Params & 12B & 6B & 8B & 25B & 9B & 12B & 72B & 9B & 7B & 7B & 70B & \\
Acc (\%) & 6.0 & 5.4 & 16.1 & 7.7 & 25.6 & 25.0 & 31.0 & 0.6 & 0.3 & 1.2 & 6.8 & \\
Ego (\%) & 86.0 & 87.5 & 39.6 & 82.7 & 17.6 & 50.0 & 24.7 & 99.4 & 99.4 & 97.0 & 76.2 & \\
\bottomrule
\end{tabular}
\caption{Summary of 35 model families evaluated on \datasetname{} (N=103 models). For each family: number of models evaluated, parameter range, average accuracy, and average egocentric response rate. Models sorted by count within each family.}
\label{tab:model_families}
\end{table*}

\subsection{Control Experiments}

To systematically dissociate the cognitive mechanisms underlying L2 VPT, we designed control experiments that isolate three distinct components: Theory of Mind (ToM), Mental Rotation (MR), and L2 VPT itself. As illustrated in \autoref{fig:task_comparison}, these tasks employ identical visual stimuli but vary cognitive demands through targeted prompts.

\textbf{ToM} isolates basic social awareness—recognizing that another agent's view differs from one's own—through a simple visibility judgment requiring only visual comparison without spatial transformation. We use "ToM" here as a simplified designation rather than implying full human theory of mind reasoning. \textbf{MR} isolates pure geometric transformation, requiring spatial reasoning without any perspective taking component. \textbf{L2 VPT} integrates both: models must adopt the monkey's perspective and mentally transform spatial representations, constituting the composite ability of genuine L2 perspective taking. By systematically varying cognitive demands while holding visual stimuli constant, we isolate each component's contribution and examine their interactions.

\subsection{Model Evaluations}

For the main experiment, we evaluate 103 publicly available VLMs spanning a wide range of model families, parameter sizes (1B to 90B), and architectures. For the control experiments, we evaluate a subset of 24 models across all three tasks (ToM, MR, and L2 VPT) to enable systematic comparison of cognitive components.

All models are evaluated under zero-shot conditions with no fine-tuning or in-context demonstrations. We invoke each model using its native API or open-source implementation, preserving default inference behavior. To standardize evaluation across heterogeneous model types, we developed a lightweight toolkit that formats prompts, invokes inference, parses responses, and maps model outputs to answer categories (Correct, Egocentric, Confusable, Random, Fail) based on ground-truth 180° rotations. All models receive identical image-text inputs. Our evaluation encompasses both direct-response models and those with Chain-of-Thought capabilities, enabling comprehensive behavioral analysis across different reasoning modes while maintaining experimental control.

\section{Results}
\label{sec:results}

\subsection{Main Results: Egocentric Bias}
\begin{figure}[h]
  \centering
  \includegraphics[width=\linewidth]{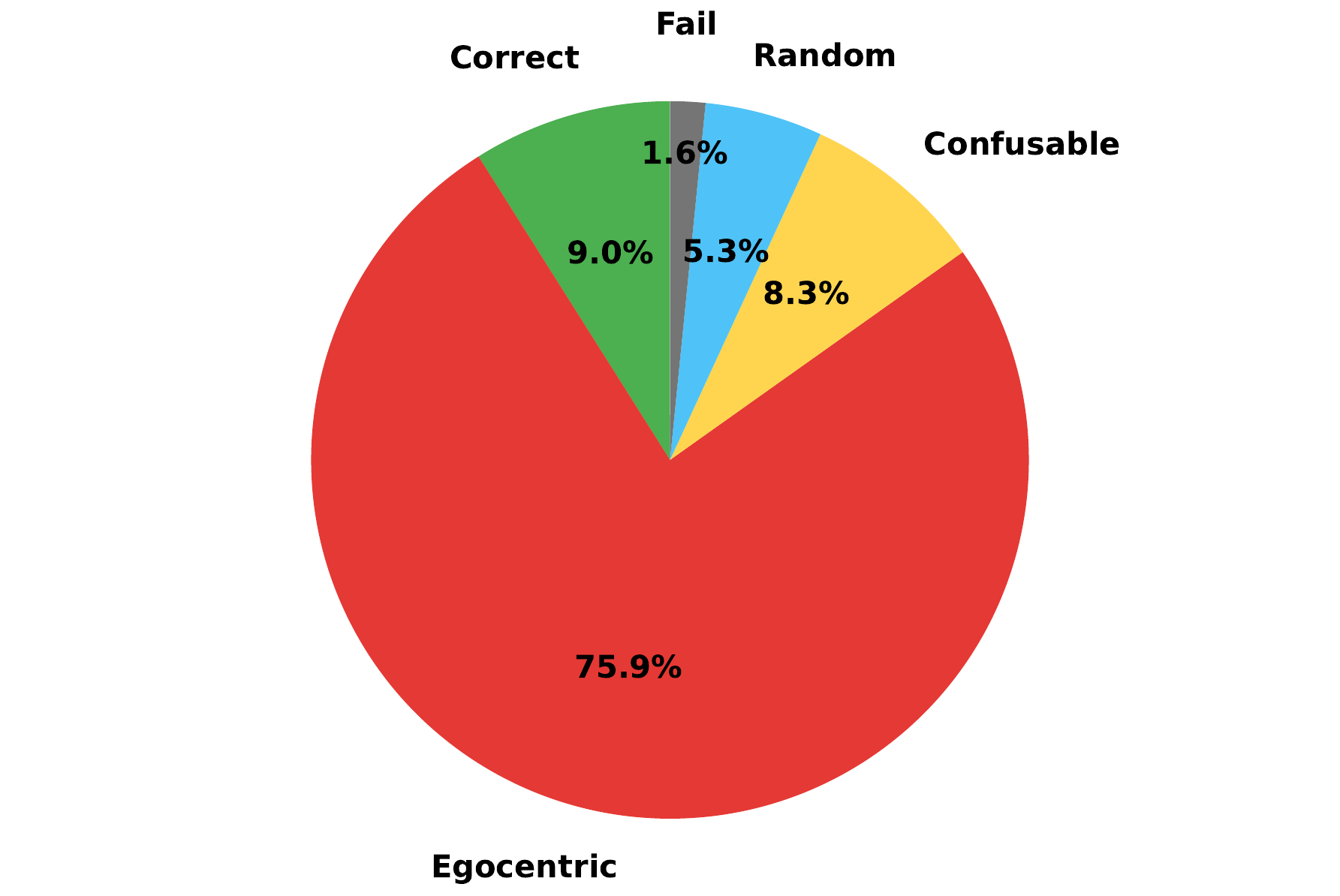}
  \caption{Error type distribution across 103 models on L2 visual perspective taking tasks. Categories include Correct, Egocentric, Confusable, Random, and Fail responses. Complete results by model families are presented in Table \ref{tab:model_families}.}
  \label{fig:error_types}

  \vspace{-2mm}
  
\end{figure}

We begin by evaluating overall model accuracy on \datasetname. Each item is formatted as a four-way multiple-choice question, yielding a 25\% chance-level baseline. However, as illustrated in \autoref{fig:error_types}, 91.3\% of models perform substantially below this threshold. Average accuracy across all 103 models is 8.96\%, with the best-performing model achieving only 33.9\% and median performance at 5.36\%—well below chance. These results reveal not merely poor performance but systematic failure: current VLMs are not solving the task via structured spatial reasoning from the agent's viewpoint, but instead rely on superficial heuristics that bypass spatial transformation altogether.

\begin{figure*}[t]
  \centering
  \includegraphics[width=\textwidth]{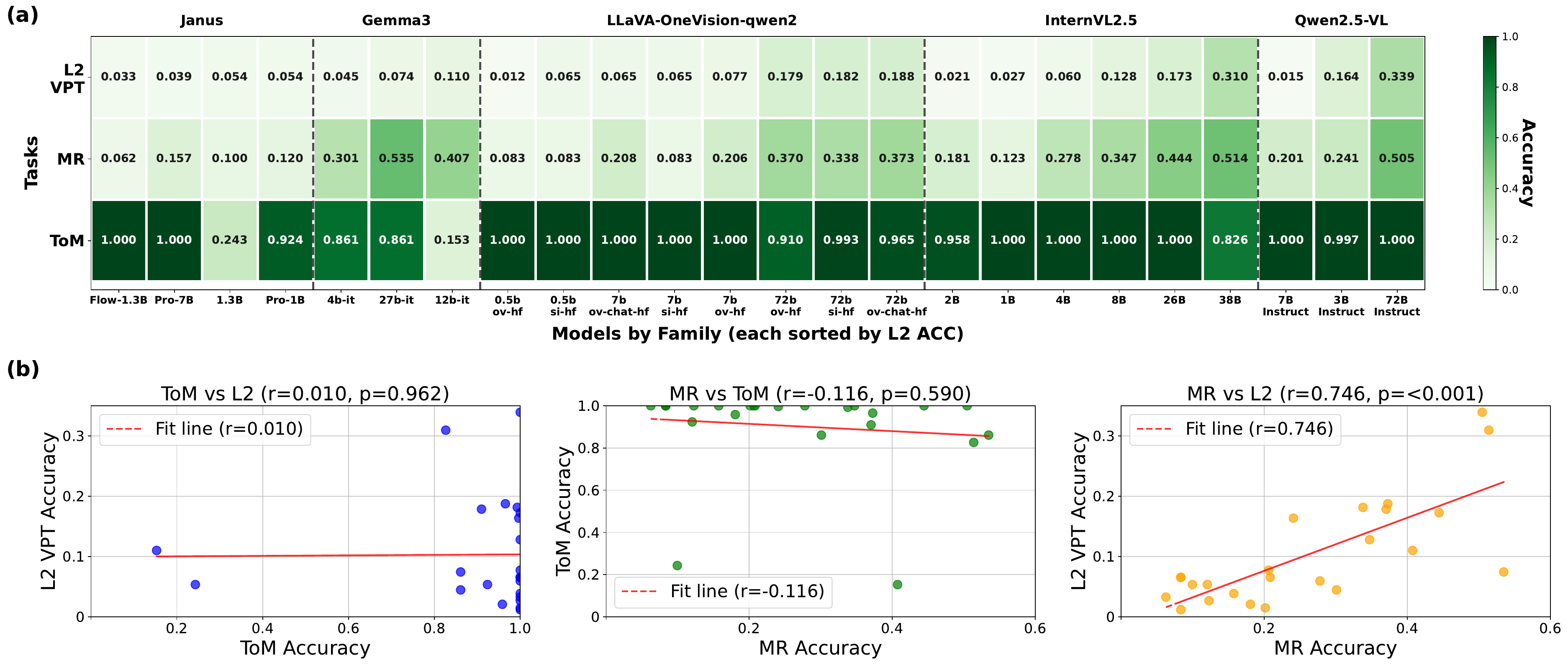}
  \caption{\textbf{Control Experiment Results: Model Performance and Task Correlations.}
  \textbf{(a)} Performance comparison of 24 models on three cognitive tasks: ToM (theory of mind), L2 (Level 2 perspective taking), and MR (mental rotation). Models are grouped by family and sorted by L2 accuracy within each family.
  \textbf{(b)} Correlation analysis between the three cognitive tasks, showing the relationships between ToM, L2 VPT, and MR performance across the 24 models.}

  \label{fig:control_experiment_results}
\end{figure*}
The error distribution reveals the mechanism underlying this failure. Egocentric responses—where models reproduce the string visible from the camera's viewpoint while disregarding the monkey's perspective—account for 75.88\% of all answers. In contrast, correct answers comprise only 8.96\%, while confusable (8.29\%), random (5.30\%), and fail (1.57\%) responses remain comparatively rare. This pattern provides diagnostic evidence: if models were attempting mental rotation but failing at visual recognition, we would expect higher rates of confusable errors. Instead, the overwhelming dominance of egocentric responses indicates that models default to their own visual reference frame without engaging in spatial transformation. This "\textbf{egocentric bias}" persists regardless of model architecture or training approach.

To examine whether this bias extends to models' reasoning processes, we analyzed representative Chain-of-Thought traces from models that generate explicit reasoning steps. Qualitative analysis reveals that even when models articulate reasoning, they frequently invoke egocentric assumptions or superficial visual heuristics rather than accurate mental rotation. Representative examples of these failure modes—including egocentric rationalization, contour-based substitution, and partial transformation errors—are detailed in~\ref{app:cot_examples}. These findings demonstrate that egocentric bias persists even under structured reasoning approaches.


\subsection{Control Results}

\label{sec:cognitive_separation}

The control experiment reveals the cognitive mechanisms underlying models' egocentric bias. We evaluated 24 models across three tasks (ToM, MR, L2 VPT) to dissociate component abilities. As shown in \autoref{fig:control_experiment_results}a, most models exhibit a consistent performance hierarchy: ToM $>$ MR $>$ L2 VPT.

Models demonstrate excellent ToM performance (mean: 90.4\%), indicating they recognize that agents positioned differently see different things. However, MR performance remains modest (mean: 26.1\%, barely above the 25\% chance baseline), and L2 VPT performance is markedly lower (mean: 10.3\%). Notably, architectural specialization emerges: LLaVA-OneVision models excel at ToM (mean: 98.4\%) but struggle with MR (mean: 21.8\%), while Gemma models show weaker ToM (mean: 62.5\%) but stronger MR (mean: 41.4\%).

Correlation analysis across the 24 models (\autoref{fig:control_experiment_results}b) reveals critical cognitive dissociations. ToM and L2 VPT show no correlation ($r = 0.010$, $p = 0.962$), as do MR and ToM ($r = -0.116$, $p = 0.590$), indicating these are independent processes. However, MR and L2 VPT correlate strongly ($r = 0.746$, $p < 0.001$), confirming that mental rotation is necessary for L2 perspective taking.

Most critically, we observe a systematic \textbf{compositional deficit}: L2 VPT performance consistently falls below what would be expected from models' component abilities. If models could successfully integrate ToM and MR, L2 VPT accuracy should approximate their product (ToM $\times$ MR). However, examining individual models reveals this is not the case. For instance, Qwen2.5-VL-72B-Instruct achieves perfect ToM (1.000) and above-chance MR (0.505), yet its L2 VPT performance (0.339) falls substantially below the expected 0.505. Similarly, InternVL2.5-38B demonstrates ToM = 1.000 and MR = 0.514, but achieves only 0.310 on L2 VPT—a 40\% deficit from the expected 0.514. This pattern holds across model families: 22 of 24 models (91.7\%) show L2 VPT performance below the factorial prediction, with a median deficit of 52.3\% relative to expected performance. Strikingly, 16 models (66.7\%) achieve above-chance MR performance ($>$0.25) while remaining at or below chance on L2 VPT ($\leq$0.25), despite near-perfect ToM scores.

This dissociation reveals that models' L2 VPT failures stem not merely from deficient mental rotation or social awareness in isolation, but from a fundamental inability to \textbf{integrate these capacities within a situated reasoning context}. Models possess the constituent abilities—recognizing different viewpoints (ToM) and performing geometric transformations (MR)—yet systematically fail to deploy them jointly when the task demands coordinated perspective taking and spatial transformation.

\section{Discussion}

Our findings reveal systematic egocentric bias in VLMs' L2 visual perspective taking: models overwhelmingly reproduce the camera viewpoint rather than simulating the agent's perspective. Our systematic answer choice design—distinguishing correct, egocentric, confusable, and random responses—enables fine-grained behavioral analysis. Through controlled experiments, we demonstrate this reflects not only failures in mental rotation but also a compositional deficit: models fail to integrate social awareness with spatial reasoning capabilities in situated contexts. This behavioral reductionistic approach, inspired by developmental psychology \citep{piaget1977development, flavell2013perspectives, zhao2016nine}, dissociates component processes to expose the cognitive architecture underlying failures.

The observed pattern echoes aspects of Piaget's account of childhood egocentrism as a signature of the preoperational stage \citep{piaget1954construction}: an inability to coordinate one's own perspective with another's through structured mental operations. Our control experiments reveal an analogous limitation in VLMs—models recognize that agents see differently (high ToM accuracy) yet fail when this awareness must be operationalized through spatial transformation (low L2 VPT despite above-chance MR). Most models show L2 VPT performance below what their component abilities predict, suggesting they possess cognitive building blocks but lack integrative mechanisms to deploy them in context. This illuminates a fundamental question for AI development: does complex perspective taking emerge compositionally from scaling simpler capacities? Our findings indicate that current VLMs, in addition to improvements in spatial reasoning, may require targeted mechanisms for binding social awareness to spatial operations.

Our results illuminate broader concerns regarding the architectural constraints of VLMs. Humans perform L2 VPT through model-based spatial reasoning supported by mental simulations—constructing internal scene models and mentally simulating transformations \citep{shepard1971mental, zacks2008neuroimaging, gunia2021brain}. Recent work argues that VLMs rely on coarse-grained visual encodings that may be mechanistically unsuitable for fine-grained spatial reasoning \citep{zhang2024exploring, luo2025rethinking, fu2025hidden}. Our findings align with this view: the compositional deficit we observe—where models fail to integrate ToM and MR despite possessing both—may reflect VLMs' reliance on learned visual-linguistic associations rather than structured spatial representations that support systematic transformations. This explains why Chain-of-Thought prompting fails to improve L2 VPT \citep{li2025core, wang2025vision}: linguistic reasoning operates disconnected from spatial structure, producing fluent but spatially invalid rationalizations that reinforce egocentric biases. These limitations resonate with broader concerns about foundation models' inability to perform structured physical reasoning \citep{wang2024can, xu2024hallucination, sun2024probing, sun2025probing, gao2025vision}.

Addressing these limitations calls interventions at multiple levels: targeted trainings on multi-view or egocentric-to-allocentric data to encourage perspective-invariant representations \citep{luo2025philosophical, yin2025spatial}; systems supporting model-based simulation beyond purely pattern-based retrieval; and even novel architectures that enable fine-grained visual encoding or explicit 3D scene representations to provide structured spatial substrates. Diagnostic benchmarks like \datasetname{} that isolate cognitive components will be essential for tracking progress.

\section{Conclusion}

We introduced \datasetname, a diagnostic benchmark for Level-2 visual perspective taking that isolates 180° rotation from another agent's viewpoint. Evaluating 103 VLMs reveals systematic egocentric bias: models overwhelmingly reproduce the camera viewpoint, with Chain-of-Thought reasoning often amplifying rather than mitigating this failure. Our controlled experiments expose the cognitive architecture underlying this bias. Models demonstrate high social awareness (ToM: 90.4\%) yet perform near chance on mental rotation (MR: 26.1\%) and catastrophically on L2 VPT (10.3\%). Critically, L2 VPT performance falls below what component abilities predict, revealing a compositional deficit—models possess cognitive building blocks but cannot integrate them in situated contexts. This echoes findings from developmental psychology that visual perspective taking requires coordinating awareness with structured spatial operations, highlighting that current VLMs need architectural innovations supporting model-based spatial reasoning rather than purely pattern-matching mechanisms.

\section*{Acknowledgments}
This work was supported by a MiraclePlus Compute Grant to Growing AI Like A Child (\url{https://growing-ai-like-a-child.github.io/}). We thank Tomer Ullman for inspiration for the work, and anonymous reviewers for their feedback.

\printbibliography

\appendix

\renewcommand{\thefigure}{A\arabic{figure}} 
\renewcommand{\thetable}{A\arabic{table}} 
\setcounter{figure}{0}
\setcounter{table}{0}

\clearpage

\noindent{\Large\textbf{Appendix}}
\vspace{0.3em}

\renewcommand{\thesection}{Appendix \Alph{section}}
\setcounter{section}{0}

\vspace{-2mm}

\section{Counterbalancing Design and Validation}\label{app:counterbalancing}

To eliminate potential bias from fixed answer positions, we systematically permute the four choices across 12 counterbalanced layouts per item. This design ensures that each answer type appears equally often in each position (A, B, C, D), thus counterbalancing any positional bias.

\begin{table}[ht]
\centering\small
\begin{tabular}{ccccc}
\toprule
Cond. & A & B & C & D \\
\midrule
1  & 18 & 78 & 81 & 87 \\
2  & 18 & 81 & 78 & 87 \\
3  & 18 & 81 & 87 & 78 \\
4  & 81 & 18 & 78 & 87 \\
5  & 81 & 18 & 87 & 78 \\
6  & 81 & 87 & 18 & 78 \\
7  & 78 & 18 & 81 & 87 \\
8  & 78 & 81 & 18 & 87 \\
9  & 78 & 81 & 87 & 18 \\
10 & 81 & 78 & 18 & 87 \\
11 & 81 & 78 & 87 & 18 \\
12 & 81 & 87 & 78 & 18 \\
\bottomrule
\end{tabular}
\caption{Twelve counter-balanced answer permutations used in the experiment. Each condition defines a unique mapping between answer options A–D and four possible answer types: \textbf{18} = Correct, \textbf{81} = Egocentric, \textbf{78} = Confusable, and \textbf{87} = Random. This design ensures that each answer type appears equally often in each position, thus counterbalancing any positional bias.}
\label{tab:answer_permutations}
\end{table}

\begin{figure}[h]
\centering

\vspace{-8mm}
\includegraphics[width=0.9\linewidth]{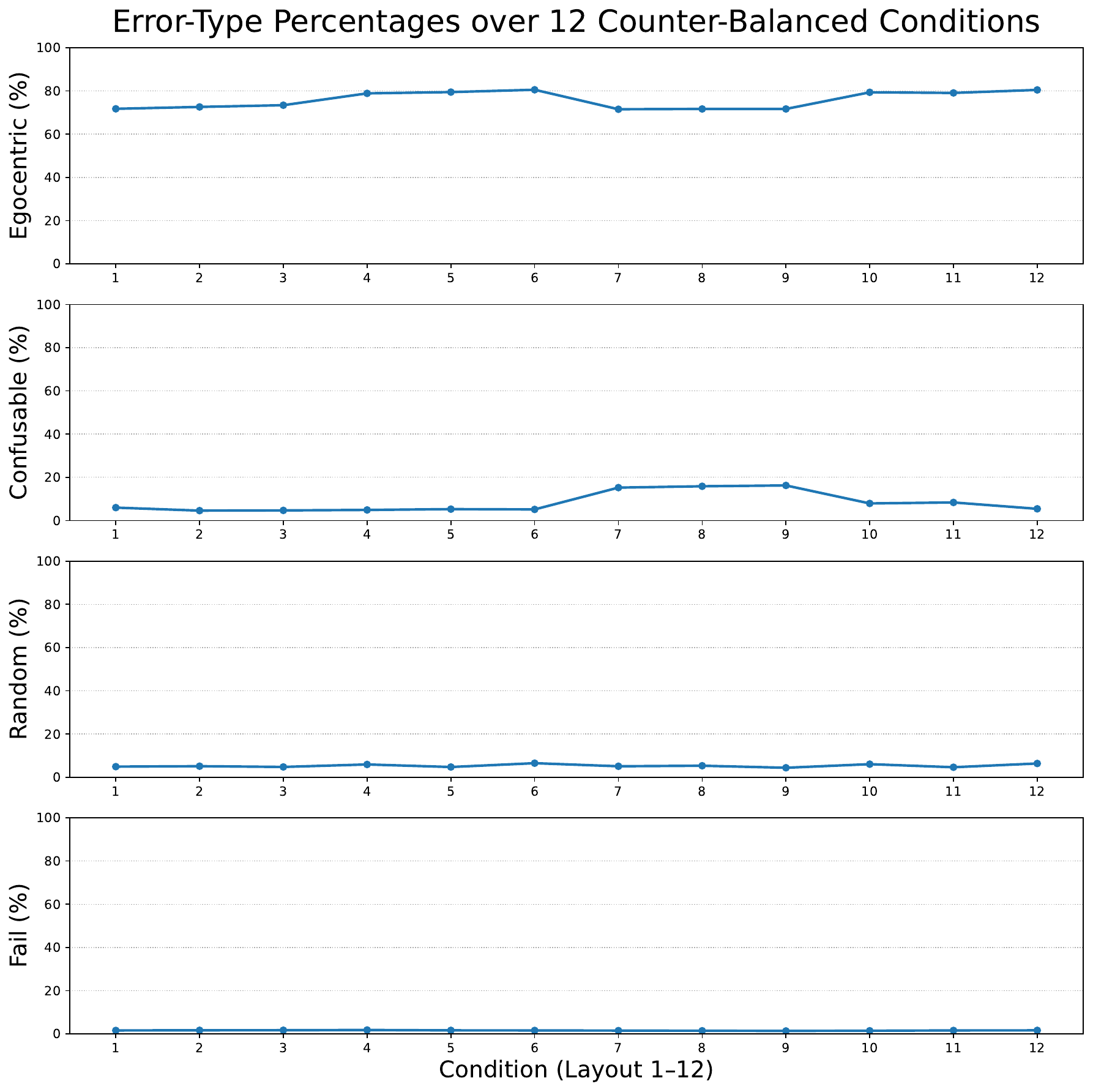}

\caption{Error curves across 12 answer-position permutations. Egocentric error rates fluctuate by $\approx$9pp, while confusable error rates fluctuate by $\approx$12pp, with particularly high confusable rates in layouts 7–9; overall accuracy shifts remain within $\approx$11pp. This suggests that answer position has limited influence on model behavior.}
\label{fig:error_curves_per_condition}
\end{figure}

Analysis of error patterns across different layouts confirms that positional effects are minimal, as shown in \autoref{fig:error_curves_per_condition}.

\section{Qualitative Analysis of Model Reasoning}\label{app:cot_examples}

To better understand the cognitive processes underlying the observed error patterns, we collected free-form outputs from representative models that generate Chain-of-Thought reasoning. Our analysis focuses on three evaluated models—deepseek-vl2, DeepSeek-VL-7B-Chat, and OpenFlamingo-9B—on representative \datasetname items. Below are illustrative examples of the most common failure patterns: egocentric rationalizations, contour-based substitutions, and partial transformation errors.

\medskip
\paragraph{Example A: Egocentric Rationalization}
\textbf{Input:} Card shows "81". \\
\textbf{Prompt:} \textit{What does the monkey see on the card? Choose A, B, C, or D.} \\
\textbf{Options:} A. 18 \quad B. 78 \quad C. 81 \quad D. 87 \\
\textbf{Model (deepseek-vl2):}
\begin{quote}
The number on the card is clearly 81. The monkey is looking at the back of the card, but the number doesn't change. Therefore, the answer is C.
\end{quote}
\textbf{Ground truth:} A (18) \quad \textbf{Error type:} Egocentric

\vspace{-2mm}

\medskip
\paragraph{Example B: Contour Substitution Confusion}
\textbf{Input:} Card shows "q". \\
\textbf{Prompt:} \textit{What does the monkey see on the card? Choose A, B, C, or D.} \\
\textbf{Options:} A. b \quad B. d \quad C. p \quad D. q \\
\textbf{Model (DeepSeek-VL-7B-Chat):}
\begin{quote}
When the monkey looks at the card from the opposite side, the letter 'q' will look like a 'p'. So I choose C.
\end{quote}
\textbf{Ground truth:} A (b) \quad \textbf{Error type:} Confusable

\vspace{-2mm}

\medskip
\paragraph{Example C: Partially Correct CoT with Wrong Conclusion}
\textbf{Input:} Card shows "nod". \\
\textbf{Prompt:} \textit{What does the monkey see on the card? Choose A, B, C, or D.} \\
\textbf{Options:} A. pou \quad B. uop \quad C. don \quad D. nod \\
\textbf{Model (OpenFlamingo-9B):}
\begin{quote}
To simulate the monkey's view, I need to reverse and rotate the characters. 'n' becomes 'u', 'o' stays the same, and 'd' becomes 'p'. So the answer is B: 'uop'.
\end{quote}
\textbf{Ground truth:} A (pou) \quad \textbf{Error type:} Confusable (wrong character order)

\medskip
These examples illustrate that even when models engage in explicit reasoning steps, they frequently default to egocentric assumptions or rely on superficial visual heuristics rather than performing accurate mental rotation. The qualitative analysis reveals that the observed egocentric bias persists even when models attempt structured reasoning approaches, suggesting that the fundamental limitation lies not in the reasoning format but in the underlying spatial cognitive mechanisms.




\end{document}